\pdfoutput=1
\documentclass[11pt]{article}

\usepackage[]{EMNLP2023}

\usepackage{times}
\usepackage{latexsym}

\usepackage[T1]{fontenc}
\usepackage[utf8]{inputenc}

\usepackage{microtype}

\usepackage{inconsolata}

\usepackage{graphicx}
\usepackage{subcaption}
\usepackage{dblfloatfix} 

\usepackage{algorithm}
\usepackage{listings}
\usepackage{xcolor}

\definecolor{codegray}{rgb}{0.5,0.5,0.5}
\definecolor{codepurple}{rgb}{0.58,0,0.82}

\lstdefinestyle{mystyle}{
    commentstyle=\color{codegray},
    keywordstyle = \color{blue},
    numberstyle=\tiny\color{codegray},
    stringstyle=\color{codegreen},
    basicstyle=\ttfamily\footnotesize,
    breakatwhitespace=false,         
    breaklines=true,                 
    captionpos=b,                    
    keepspaces=true,                 
    numbers=left,                    
    numbersep=5pt,                  
    showspaces=false,                
    showstringspaces=false,
    showtabs=false,                  
    tabsize=1
}

\usepackage{booktabs}

\title{Optimizing Hand Region Detection in MediaPipe Holistic Full-Body Pose Estimation to Improve Accuracy and Avoid Downstream Errors}

\author{Amit Moryossef \\
  \url{sign.mt} \\
  University of Zurich \\
  \texttt{amit@sign.mt}}

\begin{document}
\maketitle

\begin{abstract}
This paper addresses a critical flaw in MediaPipe Holistic's hand Region of Interest (ROI) prediction, which struggles with non-ideal hand orientations, affecting sign language recognition accuracy. We propose a data-driven approach to enhance ROI estimation, leveraging an enriched feature set including additional hand keypoints and the z-dimension. Our results demonstrate better estimates, with higher Intersection-over-Union compared to the current method.
Our code and optimizations are available at \url{https://github.com/sign-language-processing/mediapipe-hand-crop-fix}.
\end{abstract}

\section{Introduction}

In recent years, pose estimation \cite{pose:pishchulin2012articulated,pose:cao2018openpose,pose:alp2018densepose} has emerged as a fundamental component for various applications ranging from action recognition and interactive gaming to the more nuanced field of sign language processing. Among the tools at the forefront of this technological revolution, MediaPipe Holistic \cite{mediapipe2020holistic} distinguishes itself by offering real-time processing capabilities across a diverse array of devices, coupled with its adaptability in different runtime environments. This flexibility has facilitated its adoption across both research and practical applications.

However, the MediaPipe Holistic approach exhibits a significant flaw. The heuristic they use\footnote{\url{https://github.com/google/mediapipe/blob/master/mediapipe/modules/holistic_landmark/hand_landmarks_from_pose_to_recrop_roi.pbtxt}} for determining the region of interest (ROI) for hands was designed for scenarios where the hand's plane is parallel to the camera—--a design choice that does not hold in numerous practical situations. This limitation can lead to inaccuracies in hand ROI prediction, which subsequently affect the detection of hand keypoints, and compromises the overall accuracy of the full-body pose estimation \cite{moryossef2021evaluating}.

The naiveté of the existing hand ROI prediction method in MediaPipe Holistic typically manifests when dealing with non-ideal hand orientations. Given the applications of MediaPipe Holistic in domains such as sign language recognition, enhancing the robustness of hand ROI prediction is needed for accurate downstream solutions.

Concretely, their approach (Algorithm \ref{alg:current}) extracts the body pose using BlazePose \cite{bazarevsky2020blazepose}, which includes four hand keypoints per hand - the \emph{Wrist}, the \emph{Index MCP}, the \emph{Pinky MCP}, and the \emph{Thumb}. 
Then, a rough ROI crop estimate is calculated from the first three points by estimating the position of the center of the hand from the \emph{Index MCP} and \emph{Pinky MCP}, and calculating the hand size as double the distance from the center to the \emph{Wrist}. The center is then shifted, and a bounding box around it is estimated by scaling the hand size using hard-coded values. 
The 2D hand orientation is then estimated from the angle between the wrist and the center, and a rough crop is produced. This rough crop is fed to a re-cropping model which refines the bounding box, to create the hand crop. The hand crop is fed to the hand landmark model, predicting the keypoints for each hand independently.

\begin{figure*}
    \centering
    \includegraphics[width=\linewidth]{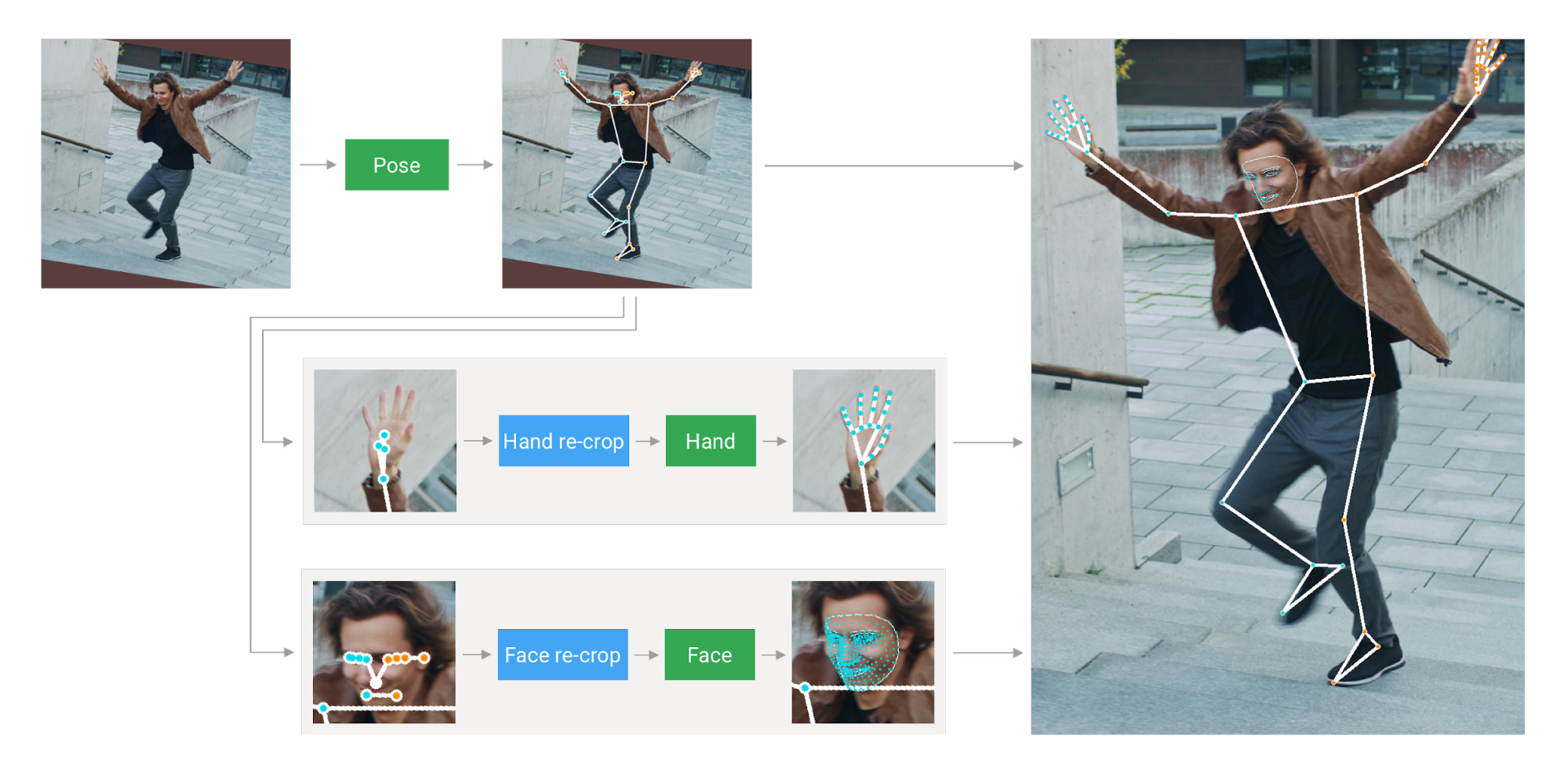}
    \caption{MediaPipe Holistic Pipeline Overview \cite{mediapipe2020holistic}.}
    \label{fig:full-pipeline}
\end{figure*}
\begin{table*}[b]
\centering
\begin{tabular}{@{}ccccc@{}}
\toprule
\textbf{Worst} & \textbf{Bad} & \textbf{OKish} & \textbf{Good} & \textbf{Best} \\ \midrule
\includegraphics[width=0.18\textwidth]{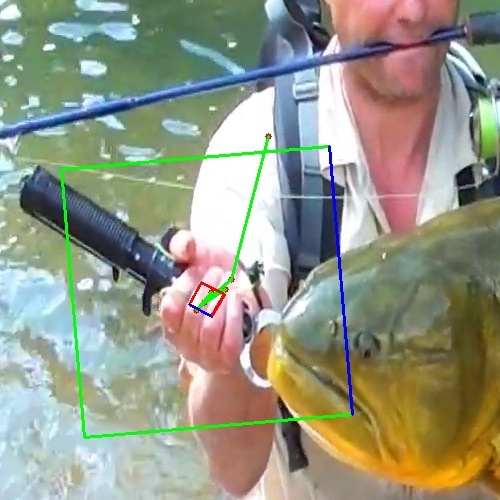} &
\includegraphics[width=0.18\textwidth]{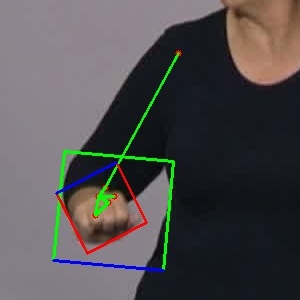} &
\includegraphics[width=0.18\textwidth]{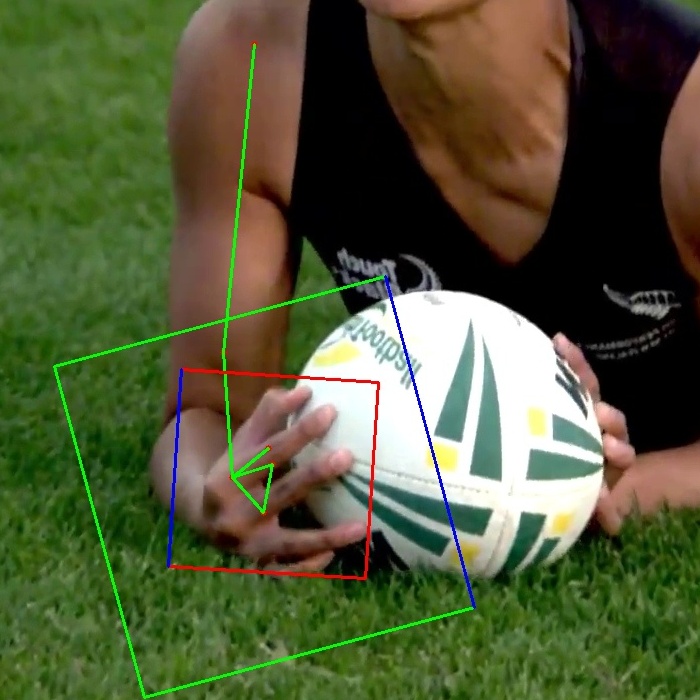} &
\includegraphics[width=0.18\textwidth]{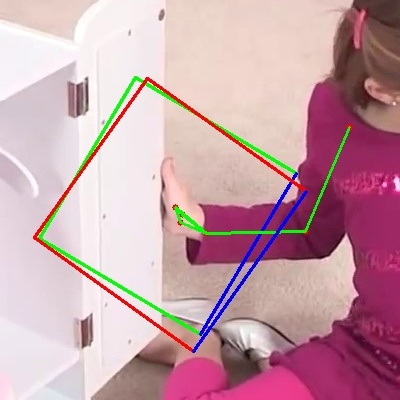} &
\includegraphics[width=0.18\textwidth]{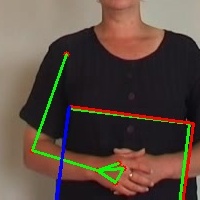} \\ \bottomrule
\end{tabular}
\caption{Selected examples of hand keypoint detections with different ROI coverages. Each image shows hand keypoints in green lines, and two bounding boxes: green for the gold ROI and red for the predicted ROI. A blue line indicates the orientation of the bounding box on the bottom edge.}
\label{table:examples}
\end{table*}
\begin{algorithm}
\caption{Hand Landmarks from Pose to ROI}
\label{alg:current}
\begin{lstlisting}[language=Python]
def calc_hand_roi(wrist, index, pinky):
    # Estimate middle finger position
    center = (2 * index + pinky) / 3
    # Estimate hand size
    size = 2 * distance(center, wrist)
    # Estimate hand 2D rotation
    rotation = angle(wrist, center) + 90
    # Shift center Y position
    center = shift(center, rotation,
                    by=(0, -0.1 * size))
    # Scale prediction
    size = 2.7 * size
    return center, size, rotation
\end{lstlisting}
\end{algorithm}

\newpage

\section{Dataset}

We utilize the Panoptic Hand DB dataset \cite{simon2017hand}\footnote{\url{https://domedb.perception.cs.cmu.edu/}}, which contains manually annotated 2D hand poses, with 1912 annotations in the training set and 846 in the testing set. Annotations encompass both \emph{right} and \emph{left} hand data, providing a comprehensive basis for analysis.

To estimate the quality of MediaPipe's ROI estimation, we first adopt the MediaPipe framework to define an ROI from the hand keypoints. This process involves bounding all keypoints and adjusting the bounding box by scaling and rotating it based on the angle between the wrist and middle finger. Then, MediaPipe Holistic is employed to predict the body keypoints. We extract the keypoints for the specific hand being analyzed (shoulder, elbow, wrist, thumb, index finger, and pinky), and for consistency, left-hand keypoints are mirrored to simulate right-handed orientation. The ROI is then determined using Algorithm \ref{alg:current}.

Table \ref{table:examples} illustrates selected instances from the dataset with varying levels of success in ROI coverage. The `Worst' and `Best' examples were automatically identified based on their coverage percentages, at $0.8$\% and $93.7$\% respectively, whereas other samples were selected manually.

\section{Methodology}
We explore an enhancement to the hand ROI calculation mechanism within the MediaPipe Holistic framework.
We recognize that the current solution only uses three hand points (\emph{wrist}, \emph{index}, and \emph{pinky}) and ignores the \emph{shoulder}, \emph{elbow}, and \emph{thumb} which can also be indicative of the hand region. Furthermore, the current solution only uses the $(x,y)$ coordinates, and ignores the predicted $z$. As we note that the model seems to perform badly when the hand is perpendicular to the camera, the $z$ dimension can be particularly useful. As history has repeatedly shown, adopting a data-driven solution is almost always a better choice \cite{SuttonBitterLesson}.

We also understand that a good solution to this problem must be very performant, and ideally as interpretable as the current solution. Since we are not members of the MediaPipe project, a complex pull-request with an additional model with obscure weights may not be accepted.

Therefore ideally, we would like to formulate a Kolmogorov-Arnold Network (KAN) \cite{liu2024kan}, as they are suitable for applications where high accuracy and interpretability are needed. Such a KAN could be formulated to use all six normalized body right-hand keypoints, alongside the image aspect-ratio ($\rho$), and predict the ROI parameters - \emph{center}, \emph{size}, and \emph{angle}.
After pruning, a KAN can be represented as simple mathematics, allowing us to deliver a solution in code, without loading additional models. 
We would start the solution by manifesting a KAN from the current mathematical solution. For example, the current \emph{size} estimation as described in Algorithm \ref{alg:current} is mathematically equivalent to the following equation for the \emph{wrist}, \emph{index} and \emph{pinky}:
$$
5.4\sqrt{\rho^2(x_1 - \frac{2x_2 + x_3}{3})^2 + (y_1 - \frac{2y_2 + y_3}{3})^2}
$$

However, due to instability in the \emph{pykan} framework, we instead use a standard MLP with two hidden layers of size of 10, to predict each of the ROI parameters separately. This delivers a fast solution of three MLPs with 332 parameters each but lacks the interpretability of a KAN.

\section{Evaluation}

To evaluate the effectiveness of our proposed methodology, we measure the improvement in hand ROI predictions. We use the following metrics:
\begin{itemize}
    \item \textbf{Center Error} - in percentage based on the image dimensions, how distance is the predicted center from the gold center.
    \item \textbf{Scale Error} - the absolute difference between the two scales, divided by the gold scale.
    \item \textbf{Rotation Error} - the absolute difference between the predicted and gold angles, circularly around 360 degrees.
    \item \textbf{IoU} - intersection-over-union of the predicted and actual hand ROIs.
\end{itemize}

We anticipate that the model will refine ROI predictions leading to higher precision in hand keypoint detection across a variety of hand orientations and movements. This should be reflected by a higher IoU.

\section{Results}

We train three separate MLPs on the train set to predict the \emph{center}, \emph{size}, and \emph{angle}.
Table \ref{table:results} shows the results of both the original and our new method on the test set. It shows that the MLP manages to better predict the \emph{center} and \emph{size} of the ROI, but fails to better predict the rotation. We believe that this is due to the simplicity of our network, containing only linear layers with \emph{relu} activations.

\begin{table}[htbp]
\centering
\resizebox{\linewidth}{!}{%
\begin{tabular}{@{}lcccc@{}}
\toprule
Method & IoU $\uparrow$ & Center $\downarrow$ & Scale $\downarrow$ & Rotation $\downarrow$ \\ \midrule
Original & 57\% & 2.51\% & 30.37\% & \textbf{32.08} \\
MLP & \textbf{63\%} & \textbf{2.15\%} & \textbf{17.91\%} & 56.96 \\
\bottomrule
\end{tabular}
}
\caption{Comparison of our method to the original.}
\label{table:results}
\end{table}

\begin{figure}[h]
    \centering
    \includegraphics[width=\linewidth]{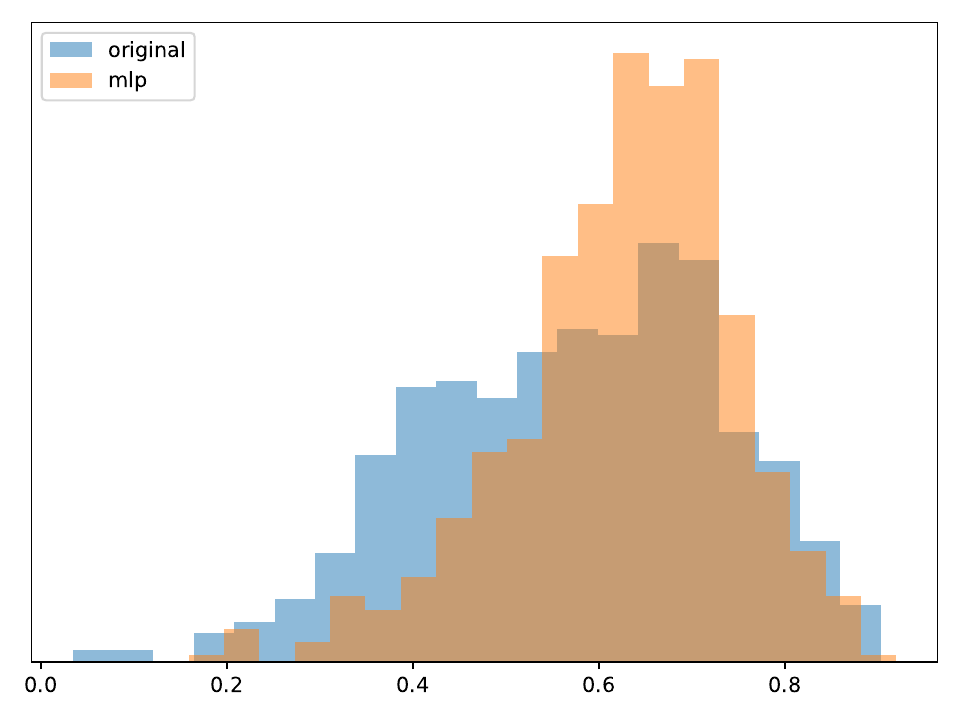}
    \caption{Histogram of IoU scores per method.}
    \label{fig:hist}
\end{figure}

Importantly, on the test set, we find that while the minimum IoU using the original method is 3\%, our new method archives a minimum of 16\%, indicating that it might work better for edge cases.
Comparing the IoU per image, the MLP wins against the original method 63\% of the time. Figure \ref{fig:hist} shows the distribution of scores per method.

Therefore, unless further optimized, we believe the final solution should use the MLP to predict the \emph{center} and \emph{scale}, and use the heuristic to calculate the rotation.

\section{Limitations}

The issues we encountered with \emph{pykan} prevented us from delivering an interpretable solution, which could prove to be better, and more easily accepted by the library maintainers.

While being an improvement on the current methodology, our solution should not be the final one. Users of MediaPipe with more time on their hands could explore additional solutions and validate them on larger amounts of data. 
Our code is available to ease these future optimizations \url{https://github.com/sign-language-processing/mediapipe-hand-crop-fix}.

\newpage

\bibliography{background,anthology,custom}
\bibliographystyle{acl_natbib}

\end{document}